\documentclass{article}

\usepackage{PRIMEarxiv}

\usepackage[numbers,sort&compress]{natbib}
\usepackage[utf8]{inputenc} 
\usepackage[T1]{fontenc}    
\usepackage{hyperref}       
\usepackage{url}            
\usepackage{booktabs}       
\usepackage{amsfonts}       
\usepackage{nicefrac}       
\usepackage{microtype}      
\usepackage{lipsum}
\usepackage{fancyhdr}       
\usepackage{graphicx}       
\graphicspath{{media/}}     

\usepackage{subcaption}
\usepackage{tabularx}
\usepackage{array}
\newcolumntype{Y}{>{\centering\arraybackslash}X}
\newcolumntype{Y}{>{\centering\arraybackslash}X}
\usepackage{xspace}
\usepackage{amsmath}
\usepackage{float}

\newcommand{\method}{TriCLE\xspace}
\newcommand{\eg}{e.g.\xspace}  
\title{TriCLE: Tri-Modal Vision-Language Reasoning for Edge-Deployed Fine-Grained Clustering
}

\author{
  Kishor Datta Gupta \\
  Department of Cyber Physical Systems \\
  Clark Atlanta University \\
  Atlanta, GA, USA \\
  \texttt{kgupta@cau.edu}
  \And
  Md. Mahfuzur Rahman \\
  Department of Cyber Physical Systems \\
  Clark Atlanta University \\
  Atlanta, GA, USA \\
  \texttt{mdmahfuzur.rahman@students.cau.edu}
  \And
  Fahad Rahman \\
  Department of Computer Science and Engineering \\
  United International University \\
  Dhaka, Bangladesh \\
  \texttt{frahman203014@bscse.uiu.ac.bd}
  \And
  Ahmed Rafi Hasan \\
  Department of Computer Science and Engineering \\
  United International University \\
  Dhaka, Bangladesh \\
  \texttt{ahasan191131@bscse.uiu.ac.bd}
  \And
  Faysal Mehrab Chowdhury \\
  Department of Electrical and Computer Engineering \\
  North South University \\
  Dhaka, Bangladesh \\
  \texttt{faysal.chowdhury4@northsouth.edu}
  \And
  Mohd Ariful Haque \\
  Department of Cyber Physical Systems \\
  Clark Atlanta University \\
  Atlanta, GA, USA \\
  \texttt{mohdariful.haque@students.cau.edu}
  \And
  Roy George \\
  Department of Cyber Physical Systems \\
  Clark Atlanta University \\
  Atlanta, GA, USA \\
  \texttt{rgeorge@cau.edu}
}

\begin{document}
\maketitle

\begin{abstract}
Edge platforms used for aerial observation must interpret aircraft imagery under limited memory, limited compute, and intermittent connectivity. This setting is difficult for standard RGB-only recognition models and general-purpose vision-language models, especially when calibrated thermal and LiDAR aircraft data are unavailable. We present TriCLE, an application-oriented tri-modal vision-language system for aircraft taxonomic grouping under edge constraints. From a single RGB aircraft image, TriCLE generates a structure-preserving FLIR-style thermal view and a pseudo-LiDAR depth projection, then fuses the aligned views with task instructions in a compact Qwen3-VL backbone. The model is aligned to an expert aircraft taxonomy based on propulsion, airframe family, size, design era, and configuration, so its outputs reflect engineering-relevant similarity rather than only surface appearance. We evaluate supervised fine-tuning, rotation-preserving SFT, and three policy-alignment strategies: GRPO, GSPO, and DAPO. Sequence-level GSPO gives the strongest validation performance, reaching 88.33\% validation accuracy and 0.91 weighted F1 on valid aircraft outputs. On a held-out aircraft test partition, GSPO achieves 78.00\% accuracy and 0.793 weighted F1 while preserving 94.00\% parseable output formatting. After 4-bit quantization and attention-memory optimization, the aligned 4B model fits an 8GB deployment target and processes each tri-modal triplet in 1.48 seconds. These results support TriCLE as a practical prototype for interpretable, edge-feasible aircraft grouping, while emphasizing the need for further validation on real aligned thermal and LiDAR sensor streams.
\end{abstract}


\section{Introduction}
\vspace{-5pt}
\label{sec:introduction}

Aircraft grouping on edge platforms is a constrained visual reasoning problem. On-board systems must operate with limited memory, limited compute, and unreliable connectivity, while still distinguishing aircraft that differ in propulsion, airframe geometry, scale, design era, or configuration. RGB imagery captures appearance and texture, but it is sensitive to viewpoint, lighting, background clutter, and markings. Thermal imagery can provide heat-related cues, while LiDAR-style geometry can expose structural shape and volume. These modalities are complementary, but aligned RGB--thermal--LiDAR aircraft data are rarely available. Vision-language models (VLMs) offer a useful interface for this setting because they can combine visual evidence with natural-language task instructions and domain knowledge. However, most general VLMs are trained mainly on RGB-centric data and are not optimized for synchronized spectral and geometric reasoning under edge-compute constraints. Existing multi-modal perception methods also often assume calibrated sensors and paired annotations, which are difficult to obtain for aircraft imagery.

We present \textbf{TriCLE}, an application-oriented tri-modal VLM system for aircraft taxonomic grouping under edge constraints. Given a single RGB aircraft image, TriCLE generates a structure-preserving FLIR-style thermal view and a pseudo-LiDAR depth projection. The three aligned views are then processed by a compact 4B VLM backbone adapted with parameter-efficient fine-tuning and policy alignment. Instead of grouping aircraft only by surface appearance, TriCLE aligns predictions with an expert taxonomy based on propulsion, airframe family, size, design era, and configuration. TriCLE is intended as a deployable prototype, not as a substitute for calibrated multi-sensor aircraft hardware. The synthetic thermal and pseudo-LiDAR views provide controlled auxiliary cues for training and evaluation when real aligned sensors are unavailable. In our experiments, sequence-level GSPO gives the strongest validation result, reaching 0.91 weighted F1 on valid aircraft outputs. On a held-out aircraft test partition, GSPO achieves 0.793 weighted F1 while maintaining 94.00\% parseable output formatting. With 4-bit quantization and attention-memory optimization, the aligned model fits an 8GB deployment target and processes each tri-modal triplet in 1.48 seconds.

\noindent\textbf{Contributions.}
This work makes three contributions: 
(i) an edge-oriented tri-modal VLM pipeline for aircraft taxonomic grouping from RGB, synthetic thermal, and pseudo-LiDAR inputs; 
(ii) a synthetic sensing procedure that produces aligned FLIR-style and depth-projection views from monocular aircraft imagery; and 
(iii) a deployment-focused evaluation of policy alignment, structured-output validity, held-out aircraft performance, quantization, latency, and failure cases.

\section{Related Work}\vspace{-5pt}
\label{sec:related}
CLIP~\cite{radford2021learning} established transferable image-text representations, and LLaVA~\cite{liu2023visual} showed that visual instruction tuning can adapt frozen visual encoders for general multimodal reasoning. Broader cross-modal systems such as ImageBind~\cite{girdhar2023imagebind} align multiple sensory inputs in a shared representation space. In applied perception, TarDAL~\cite{liu2022target} improves infrared-visible fusion, while TransFusion~\cite{bai2022transfusion} advances LiDAR-camera fusion for 3D detection. These methods demonstrate the value of non-RGB cues, but they generally rely on real calibrated sensors or paired annotations. TriCLE targets the aircraft case where such synchronized data are scarce and edge deployment is required.
Synthetic sensing provides a practical alternative when paired non-RGB data are unavailable. ControlNet~\cite{zhang2023adding} preserves spatial structure during controllable generation, pseudo-LiDAR~\cite{wang2019pseudo} converts monocular depth into point-cloud-like geometry, and DPT~\cite{ranftl2021vision} provides dense depth estimation from single images. For representation learning, DeepCluster~\cite{caron2018deep}, SwAV~\cite{caron2020unsupervised}, and SimCLR~\cite{chen2020simple} show that clustering and contrastive objectives can improve embedding structure. TriCLE combines these ideas in an application-specific system that maps aircraft engineering properties into a tri-modal VLM representation space for interpretable edge-deployable grouping.
\section{Methodology} \vspace{-5pt}
\label{sec:methodology}

The aim of \method is to execute unsupervised aircraft clustering within edge computing constraints by calibrating a compact Vision-Language Model (VLM) through reasoning-based policy optimization. Our pipeline advances through four distinct operational phases: (1) synthesizing co-registered, multi-spectral inputs from monocular RGB images to create a tri-modal dataset, (2) conducting supervised fine-tuning through SVD-anchored robust parameter-efficient fine-tuning (RPSFT), (3) aligning the SFT-trained model with reasoning-based policy optimization (GRPO, GSPO, and DAPO) utilizing structured taxonomic ground truths, and (4) deploying the aligned checkpoints to execute cross-domain, zero-shot perception in real-world environments.

\subsection{Overview of System Architecture}
\label{subsec:system_overview}

\begin{figure*}[b!]
    \centering
    
    \begin{subfigure}{\textwidth}
        \centering
        \includegraphics[width=1\textwidth, keepaspectratio=true]{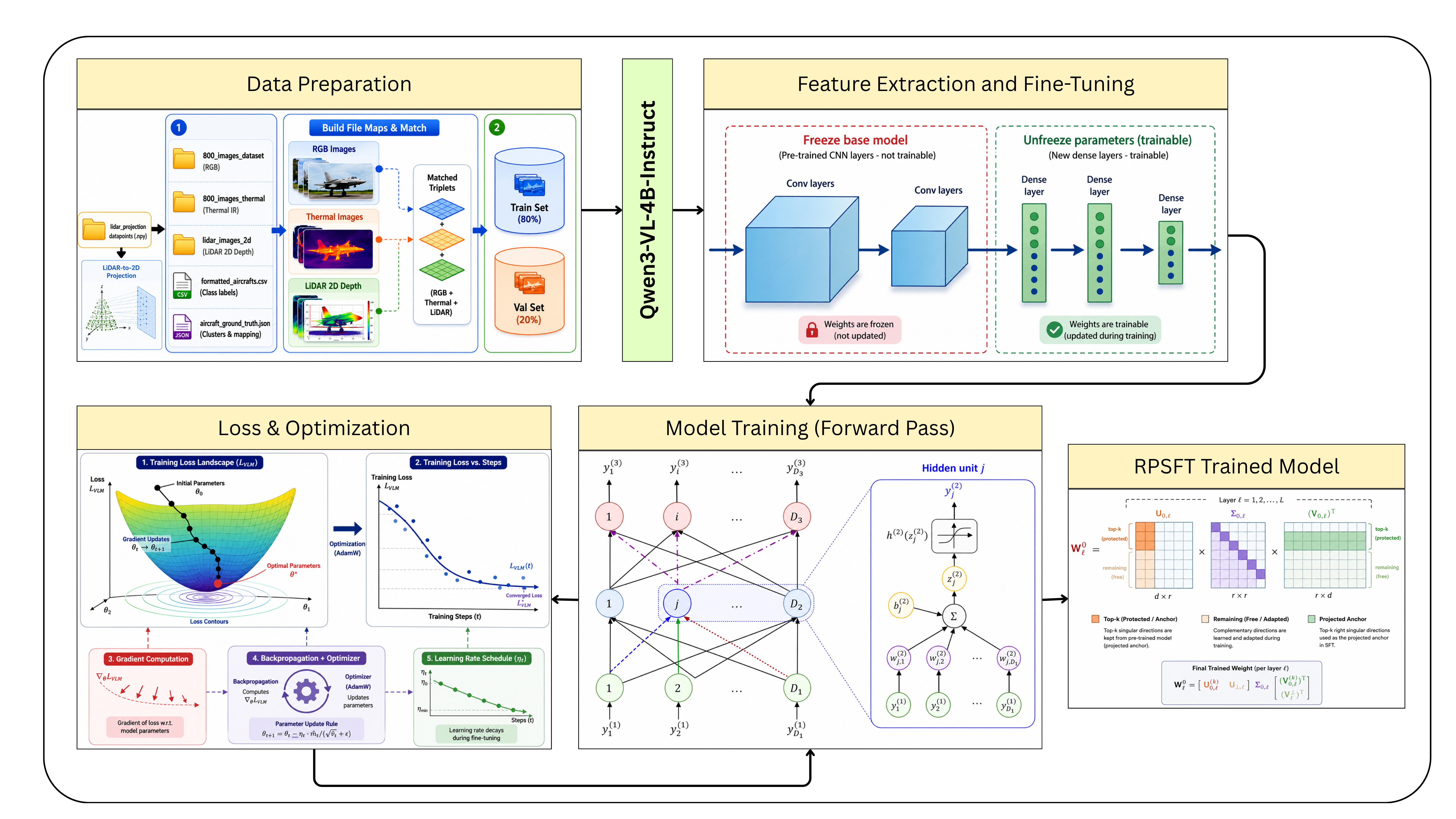}
        \caption{Data Preparation and RPSFT Pre-training}
        \label{fig:pipeline_a}
    \end{subfigure}
    
    \vspace{1em}
    
    \begin{subfigure}{\textwidth}
        \centering
        \includegraphics[width=1\textwidth, keepaspectratio=true]{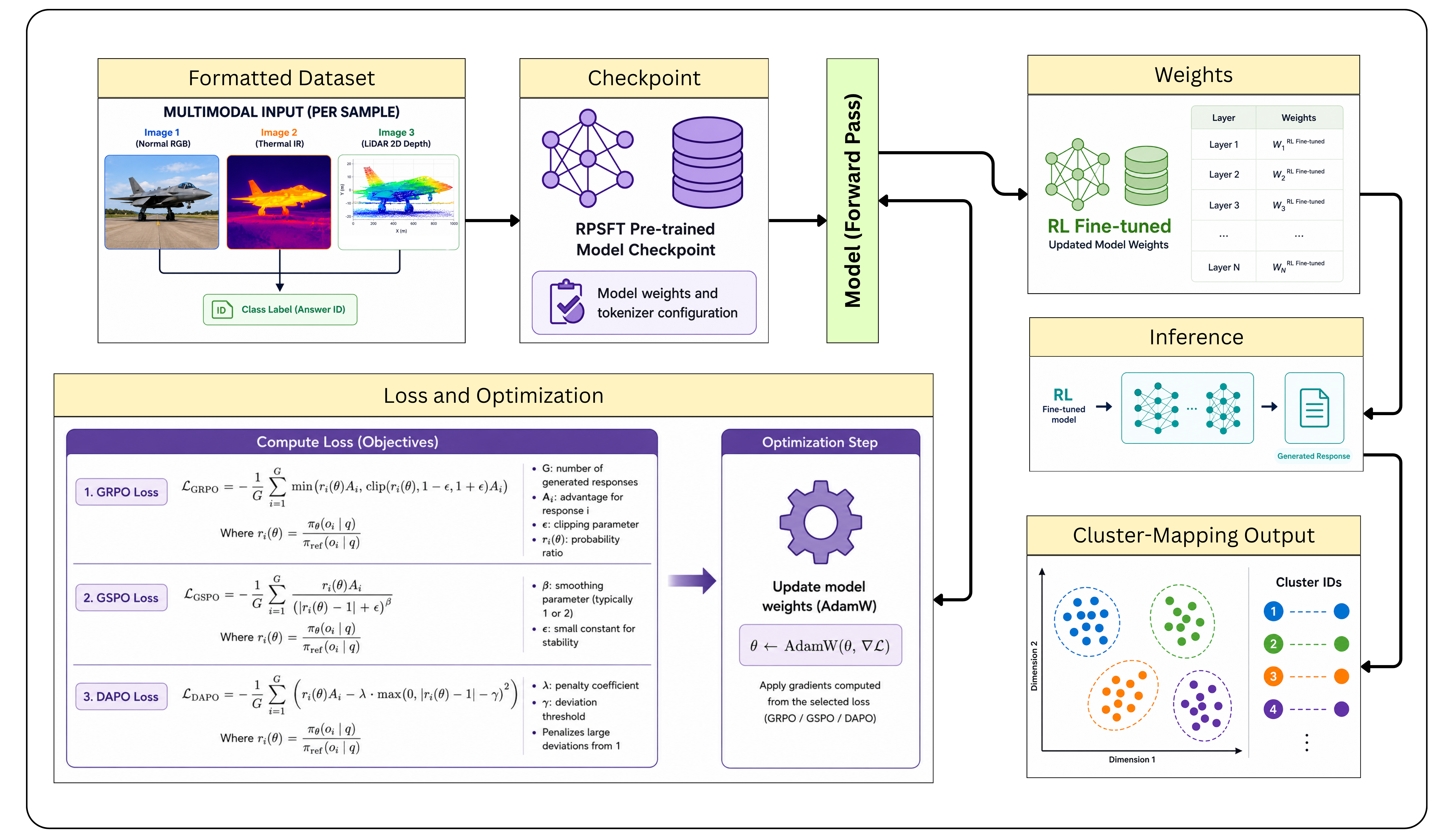}
        \caption{RL Alignment and Inference Pipeline}
        \label{fig:pipeline_b}
    \end{subfigure}
    
    \caption{The end-to-end \method system architecture.}
    \label{fig:pipeline}
\end{figure*}

The conceptual framework of \method initiates with a singular image in the visible spectrum, denoted as $I^{\text{rgb}}$. We dynamically produce a structure-preserving thermal signature $I^{\text{th}}$ and an orthographic LiDAR depth projection $I^{\text{lidar}}$. The three aligned modalities are concurrently transmitted to the attention blocks of our frozen visual encoder. The extracted visual tokens are interspersed with natural language task directives and processed via a parameter-efficient, reasoning-aligned language backbone to produce intermediate Chain-of-Thought (CoT) tokens and a final taxonomic classification. A visual summary of this execution pipeline is illustrated in Figure~\ref{fig:pipeline}.

Due to the extreme scarcity of co-registered optical, thermal, and spatial depth datasets of aircraft, we develop a tri-modal synthetic sensing pipeline that translates monocular visual inputs into aligned multi-spectral frames.

\textbf{Controllable FLIR thermal generation.}
Given an RGB aircraft image $I^{\text{rgb}} \in \mathbb{R}^{H \times W \times 3}$, we produce a synthetic long-wave infrared (LWIR) frame $I^{\text{th}} \in \mathbb{R}^{H \times W \times 3}$ that displays FLIR thermal characteristics. To maintain the original airframe geometry, wing sweep, engine location, and spatial orientation, we guide the generative process using FLUX.1-dev~\cite{blackforestlabs2024fluxdev} modified via a ControlNet adapter~\cite{zhang2023controlnet} conditioned on Canny edges:
\begin{equation}
    \mathbf{C}_{\text{edges}} = \mathrm{Canny}(I^{\text{rgb}}),
    \label{eq:canny}
\end{equation}
where $\mathbf{C}_{\text{edges}}$ functions as a spatial geometric constraint. To modify the base text-to-image transformer for the infrared spectrum while preventing catastrophic forgetting, we incorporate LoRA layers~\cite{hu2022lora} (rank $r{=}32$, $\alpha{=}32$) trained on an empirical FLIR dataset. The conditioning prompt includes a distinctive domain trigger token \texttt{milaircraft}, instructing the model to produce continuous thermal gradients concentrated around engine exhausts, nacelles, and leading edges.

\textbf{Monocular depth estimation and 3D back-projection.}
We derive a continuous relative depth map $D(u, v)$ from the produced thermal signature $I^{\text{th}}$ utilizing a Dense Prediction Transformer (DPT)~\cite{ranftl2021dpt}. Employing a pinhole camera model, we back-project each pixel $(u, v)$ into a 3D coordinate $[X, Y, Z]^T$ within the camera frame. Specifically, the depth $Z$ is calculated by scaling the estimated relative depth $D(u, v)$ with a positive scaling constant $s \in \mathbb{R}^+$, while the horizontal $X$ and vertical $Y$ coordinates are derived by scaling the pixel offsets from the principal point $(c_x, c_y)$ with the depth $Z$ and normalizing by the virtual focal lengths $f_x$ and $f_y$. Each reconstructed point $p_i = [X_i, Y_i, Z_i]^T$ is subsequently allocated a reflectance-style intensity $\mathbf{I}_i$ by normalizing the grayscale value of its corresponding coordinate in the thermal image $I^{\text{th}}$ to a $[0, 1]$ range, yielding an unstructured point cloud $\mathcal{P} = \{(X_i, Y_i, Z_i, \mathbf{I}_i)\}_{i=1}^N$. Statistical outlier removal and grid subsampling approximate the sparsity of active LiDAR scanners.

\textbf{VLM-compatible LiDAR orthographic projection.}
To convert $\mathcal{P}$ into an image-like depth projection $I^{\text{lidar}} \in \mathbb{R}^{H \times W \times 3}$ compatible with standard VLM vision encoders, we project points onto an elevation-width grid and normalize depth to an 8-bit intensity via the joint orthographic mapping:
\begin{equation}
    \begin{pmatrix} u \\[6pt] v \\[6pt] d_{\mathrm{norm}} \end{pmatrix}
    =
    \begin{pmatrix}
        \left\lfloor \dfrac{x - x_{\min}}{x_{\max} - x_{\min}} \cdot (W{-}1) \right\rceil \\[12pt]
        (H{-}1) - \left\lfloor \dfrac{z - z_{\min}}{z_{\max} - z_{\min}} \cdot (H{-}1) \right\rceil \\[12pt]
        \left\lfloor \dfrac{\mathrm{clip}(y,\, y_{\min},\, y_{\max}) - y_{\min}}{y_{\max} - y_{\min}} \cdot 255 \right\rfloor
    \end{pmatrix},
    \label{eq:lidar_projection}
\end{equation}
where $\lfloor\,\cdot\,\rceil$ denotes nearest-integer rounding, and $y_{\min}$, $y_{\max}$ are the 1st and 99th percentile depth values within the current frame. The resultant grayscale map undergoes morphological dilation using a $3{\times}3$ structuring element and is colorized via a JET colormap, with non-reflective zones designated a black sentinel $[0, 0, 0]$. This yields a synchronized tri-modal triplet $\mathcal{T} = (I^{\text{rgb}}, I^{\text{th}}, I^{\text{lidar}})$.

\subsection{Taxonomic Clustering and Representation Space}
\label{subsec:taxonomic_clustering}

Rather than classifying aircraft only based on visual characteristics, we construct a specialized aircraft taxonomy $\mathcal{C} = \{c_1, \ldots, c_{10}\}$ that organizes aircraft by their engineering, design, and propulsion properties (\eg, delta-wing supersonic airframes, turboprop transports, high-altitude UAVs). Each tri-modal triplet is input into our aligned model, and the final hidden state is projected into a 2048-dimensional embedding $z_i \in \mathbb{R}^{2048}$, subsequently $L_2$-normalized:
\begin{equation}
    \hat{z}_i = \frac{z_i}{\|z_i\|_2},
    \label{eq:l2_norm}
\end{equation}
mapping multi-sensor features onto a unit hypersphere. Unsupervised clustering is performed using $K$-Means, which reduces inertia:
\begin{equation}
    \arg\min_{\mathbf{S}} \sum_{j=1}^{K} \sum_{\hat{z}_i \in S_j} \|\hat{z}_i - \mu_j\|^2,
    \label{eq:kmeans}
\end{equation}
where $\mu_j$ represents the centroid of cluster $S_j$.

\subsection{Phase 2: Supervised Fine-Tuning via RPSFT}
\label{subsec:rpsft_stage}

Prior to reinforcement alignment, we perform a supervised fine-tuning stage to preserve the base model's out-of-distribution (OOD) reasoning capabilities via SVD-anchored Robust Parameter-Efficient Fine-Tuning (RPSFT). We freeze the base model and unfreeze only the attention projection weights ($W_q, W_v$) within the last 12 language backbone layers, casting them to FP32. For each unconstrained layer $l$, we compute the SVD of the initial pre-trained weight $W_0^l$:
\begin{equation}
    W_0^l = U^l S^l (V^l)^T,
    \label{eq:svd}
\end{equation}
and cache the top-$k$ singular vectors ($U_k^l$, $V_k^l$) with $k{=}64$. During training, the standard cross-entropy SFT loss is regularized by penalizing weight updates orthogonal to the cached pre-trained subspace:
\begin{equation}
\begin{split}
    \mathcal{L}_{\text{RPSFT}}(\theta) &= \mathcal{L}_{\text{SFT}}(\theta) \\
    &+ \lambda_{\text{RPSFT}} \sum_{l} \| W_{\text{diff}}^l - U_k^l (U_k^l)^T W_{\text{diff}}^l V_k^l (V_k^l)^T \|_F^2,
\end{split}
\label{eq:rpsft_loss}
\end{equation}
where $W_{\text{diff}}^l = W^l - W_0^l$ is the operational parameter drift and $\|\cdot\|_F^2$ is the squared Frobenius norm. This regularizer locks parameters within their initial pre-trained manifold, mitigating catastrophic forgetting.

\subsection{Phase 3: Policy Optimization}
\label{subsec:policy_optimization}

The RPSFT-trained model serves as both the base training policy and the reference model $\pi_{\mathrm{ref}}$. To align visual reasoning with the expert taxonomy $\mathcal{C}$, we evaluate three policy optimization techniques, described below:                                    

\textbf{Group Relative Policy Optimization (GRPO).}
For each tri-modal training input, the policy $\pi_\theta$ samples $G$ independent completions. Each is scored by a rule-based reward checking format correctness (XML tags) and taxonomic accuracy:
\begin{equation}
    R_i = \lambda_f R_{f,i} + \lambda_a R_{a,i}.
    \label{eq:grpo_reward}
\end{equation}
The relative advantage within the group is:
\begin{equation}
    A_i = \frac{R_i - \mu(\mathcal{R})}{\sigma(\mathcal{R}) + \epsilon},
    \label{eq:relative_adv}
\end{equation}
where $\mathcal{R} = \{R_j\}_{j=1}^{G}$, and $\epsilon$ prevents division by zero. The policy model is updated using the standard clipped surrogate objective, regularized via Kullback-Leibler (KL) divergence against the reference model $\pi_{\mathrm{ref}}$ to prevent language degradation following standard formulations~\cite{shao2024deepseekmath}, where $r_i(\theta) = \pi_{\theta}(y_i \mid x)/\pi_{\theta_{\text{old}}}(y_i \mid x)$ is the token probability ratio, $\eta$ is the clipping coefficient, and $\beta_{\text{KL}}$ is the KL regularization weight.

\textbf{Group Sequence Policy Optimization (GSPO).}
Standard GRPO can be vulnerable to semantic drift or infinite reasoning loops in long-context generation. GSPO addresses this by computing the probability ratio $\bar{r}_i(\theta)$ over the entire generated sequence $y_i$ via sequence-level importance sampling. The policy objective then uses sequence-level advantages $\bar{A}_i$ to update the policy weights and encourage globally coherent reasoning paths, preventing local token-level collapse~\cite{zheng2025gspo}.

\textbf{Direct Advantage Policy Optimization (DAPO).}
DAPO bypasses active advantage sampling by directly optimizing a pairwise preference loss over chosen responses $y^+$ (complete cross-spectral reasoning, correct taxonomy label) and rejected responses $y^-$ (incomplete, single-sensor, or incorrect reasoning) following standard formulations~\cite{yu2025dapo}, where $\Delta_{\theta} = \log \pi_{\theta}(y^+ \mid x) - \log \pi_{\theta}(y^- \mid x)$, and $\Delta_{\mathrm{ref}}$ is computed identically under $\pi_{\mathrm{ref}}$. To stabilize this alignment, the truncation mask:
\begin{equation}
    M_i = \mathbb{I}(\mathrm{len}(y_i) \leq L_{\text{max}})
    \label{eq:truncation_mask}
\end{equation}
filters overlong completions, preventing positive updates on truncated or non-converging generation loops.

\subsection{Phase 4: Cross-Domain Zero-Shot Inference}
\label{subsec:zero_shot_inference}

To evaluate generalizability, we deploy the frozen aligned model weights directly on the out-of-distribution DSERT-RoLL dataset~\cite{cho2026dsertroll}. The inference framework constructs a co-registered tri-spectral feed by matching RGB, long-wave thermal, and 3D LiDAR point clouds via filename prefix keys, then passes them interleaved into the visual-token sequence. The model outputs qualitative visual classes matched against our taxonomy and parsed numerical counters for Cars, Pedestrians, and Cycles without any task-specific weight updates.

\section{Dataset}\vspace{-5pt}
\label{sec:dataset_sec}

Our project is based on a dataset in which an object is learned and subsequently clustered. The primary consideration is that we are conducting subclass clustering. Thus, we require single-objective images, specifically those featuring objects of a singular class, such as cars, aircraft, dresses, etc. Another primary objective is to implement this project on an edge device, such as drones. We are utilizing dataset classes that align more closely with our objectives.
We demonstrate that utilizing thermal images and LiDAR data in conjunction with RGB images enables the training of a model that enhances subclass clustering during inference. We extensively searched for various datasets to fulfill our requirements; however, to our knowledge, we were unable to locate a single dataset that encompasses RGB images, thermal images, and LiDAR data for a straightforward objective classification. Consequently, we were required to generate a synthetic dataset for that objective. Datasets such as DESERT-RoLL~\cite{cho2026dsertroll}, R-LiViT~\cite{mirlach2025r}, TRansPose~\cite{yang2021transpose}, and FLIR~\cite{ligocki2021fully} either contain only RGB and LiDAR data or RGB and thermal images. Even when they include RGB, thermal, and LiDAR data, they fail to meet the criteria required for our experiment. 

\subsection{Synthetic Tri-Modal Data Generation} 
\label{sec:data_generation} 

\textbf{Source imagery.} RGB images are sourced from FGVC-Aircraft~\cite{maji2013finegrainedvisualclassificationaircraft}, as their single-class composition and sub-type annotations align with the objectives of our sub-class clustering. We use a YOLOv12 detector~\cite{yolo12} to systematically filter the images, thereby ensuring that the target object is prominently featured in each frame. This criterion is crucial for ensuring reliable structure-preserving synthesis and depth estimation, as we include only those images in which the regions containing aircraft constitute a minimum of $60\%$ of the frame. This results in a total of $800$ curated single-class images. We integrate the thermal and geometric modalities, ensuring that all three views are co-registered in terms of both viewpoint and resolution. 
\textbf{Thermal synthesis.}  Thermal generation can be conceptualized as a process of structure-preserving appearance transfer. In this framework, the infrared representation retains the scene's geometric and occlusion characteristics as observed in the RGB source, with the primary distinction lying in the method of pixel formation. We modify FLUX1-dev, a rectified-flow diffusion transformer, for the false-color FLIR domain through Low-Rank Adaptation (rank $r{=}32$, $\alpha{=}32$). In this process, we train solely the transformer pathway while keeping the text encoders fixed, utilizing a trigger token (\texttt{milaircraft}) to associate with the thermal appearance. During the inference phase, we maintain structural fidelity by utilizing a Canny-edge ControlNet~\cite{Zhang_2023_ICCV} (with a conditioning scale of $0.7$), accompanied by a fixed prompt that delineates the desired FLIR palette. The outcome is thermal imagery that exhibits a plausible appearance and is pixel-aligned with its RGB source. The comprehensive details regarding the training and sampling configurations are provided in the supplementary material. 
\textbf{Pseudo-LiDAR generation.} True LiDAR represents an active 3D measurement that cannot be derived from a single RGB frame. Consequently, we utilize a pseudo-LiDAR representation~\cite{wang2019pseudo}, which provides a co-registered depth and structure prior for the purpose of fusion. We conduct an estimation of a dense monocular depth map for each image, subsequently back-projecting it into a 3D point cloud. To enhance the representation, we impose a LiDAR-like structure by organizing the returns into scan rings, incorporating reflectance-style intensity, and then eliminating points that fall outside the specified range. Reintegrating the cloud into the camera frame results in a range image that is aligned with both its RGB and thermal counterparts. 
\textbf{Composition.}  The final dataset comprises 800 aligned RGB-thermal-pseudo-LiDAR triplets. Both synthetic modalities are derived from the RGB source and conform to structure-preserving constraints. Consequently, the three perspectives exhibit a consistent viewpoint and layout, which is crucial for the fusion.

\section{Experiments}\vspace{-5pt}
\label{sec:experiments}

\subsection{Experimental Setup and Benchmarks}
\textbf{Model Architecture and Hyperparameters:} The foundation of our pipeline is the compact Qwen3-VL-4B-Instruct vision-language model, initialized in \texttt{bfloat16} precision utilizing Scaled Dot-Product Attention (SDPA). To operate within strict edge-computing VRAM limits during tri-modal token interleaving, the visual processor restricts spatial dimensions between $128 \times 28 \times 28$ and $256 \times 28 \times 28$ pixels per frame. Our parameter-efficient fine-tuning strategy freezes the entire base architecture, isolating trainable parameters strictly to the attention projection matrices (\texttt{q\_proj}, \texttt{v\_proj}, \texttt{k\_proj}, \texttt{o\_proj}) within the final 12 transformer layers of the language backbone. Supervised optimization operates at a learning rate of $5 \times 10^{-5}$ with linear decay, while reinforcement learning alignment employs a tighter $1 \times 10^{-5}$ rate.
\textbf{Evaluation Metrics:} The framework is evaluated on a curated zero-shot evaluation partition capped at 100 synchronized tri-modal samples. The evaluations target two domains: in-distribution (ID) accuracy against our ten-class expert aircraft taxonomy, and out-of-distribution (OOD) accuracy using the ScienceQA benchmark to monitor catastrophic forgetting. The structural compliance of the model's reasoning is quantified using formatting validation rates (accurate XML tag extraction), taxonomic clustering accuracy, and F1-scores.

\subsection{Singular Subspace Regularization}
Standard supervised fine-tuning (SFT) can inadvertently rotate pre-trained weight matrices, leading to representational drift and the degradation of out-of-domain knowledge. To counteract this, Rotation-Preserving Supervised Fine-Tuning (RPSFT) is applied to the active attention projection weights. 
Prior to training, a singular value decomposition (SVD) is computed for each target weight matrix $W_0^l$:
$$W_0^l = U^l S^l (V^l)^T$$
The top $k=64$ left and right singular vectors ($U_k^l$, $V_k^l$) are cached as fixed anchors. The RPSFT objective function restricts parameter adaptation outside these dominant singular directions by minimizing the task-specific cross-entropy loss regularized by an SVD projection penalty:

\begin{equation}
\begin{split}
\mathcal{L}_{RPSFT}(\theta) &= \mathcal{L}_{SFT}(\theta) + \lambda_{RPSFT} \sum_{l \in \mathcal{M}'} \\
&\quad \left\| W_{diff}^{l} - U_k^{l}(U_k^{l})^T W_{diff}^{l} V_k^{l}(V_k^{l})^T \right\|_F^2
\end{split}
\end{equation}
where $W_{diff}^{l} = W^l - W_0^l$ represents the weight delta, $\mathcal{M}'$ denotes the regularized layers, and $\lambda_{RPSFT}$ dictates the penalty strength. 

Standard SFT ($\lambda_{RPSFT}=0.0$) achieves an ID aircraft classification accuracy of 63.75\%, outperforming the RPSFT baseline ($\lambda_{RPSFT}=0.1$) by 8.75\%. However, SFT suffers measurable catastrophic forgetting. RPSFT effectively preserves general reasoning capabilities, securing an OOD ScienceQA accuracy of 90.16\% with negligible computational overhead (a peak VRAM footprint of 17.26 GB).

\begin{table*}[h]
\centering
\caption{SFT vs. RPSFT Performance Trade-offs}
\begin{tabular}{lcccc}
\toprule
\textbf{Method} & \textbf{ID Acc.} & \textbf{OOD Acc.} & \textbf{Time (s)} & \textbf{VRAM (GB)} \\
\midrule
Standard SFT ($\lambda = 0.0$) & 63.75\% & 89.63\% & 3533.4 & 16.09 \\
RPSFT ($\lambda = 0.1$) & 55.00\% & 90.16\% & 3338.1 & 17.26 \\
\bottomrule
\end{tabular}
\label{tab:rpsft}
\end{table*}

\subsection{Quantitative Policy Optimization Analysis}
To align multi-modal outputs with the strict taxonomic formatting requirements, RPSFT serves as the reference policy ($\pi_{ref}$) for three reinforcement-based optimizations: Group Relative Policy Optimization (GRPO), Group Sequence Policy Optimization (GSPO), and Decoupled Advantage Policy Optimization (DAPO) %
. The reward mechanism ($R_{max} = 1.1$) incentivizes structural \texttt{<think>} block generation, domain-specific engineering vocabulary, and terminal classification accuracy.

Evaluated across the 116 validation samples, GSPO demonstrates superior alignment. By executing sequence-level rather than token-level probability ratio clipping, GSPO prevents local token collapse, achieving a peak validation accuracy of 88.33\% and a macro F1-score of 0.89. Conversely, DAPO introduces training instability due to its asymmetric bounds and soft overlong penalties, throttling valid formatting generation to 95.54\%.

\begin{table*}[h]
\centering
\caption{\footnotesize Policy Optimization Alignment Results ($n=116$)}
\begin{tabular}{lcccc}
\toprule
\textbf{Policy} & \textbf{Val. Acc.} & \textbf{Format Valid} & \textbf{Macro F1} & \textbf{Weighted F1} \\
\midrule
RPSFT Base & 55.00\% & 100.00\% & 0.49 & 0.53 \\
RPSFT+GRPO & 85.83\% & 98.12\% & 0.86 & 0.89 \\
\textbf{RPSFT+GSPO} & \textbf{88.33\%} & \textbf{98.00\%} & \textbf{0.89} & \textbf{0.91} \\
RPSFT+DAPO & 75.00\% & 95.54\% & 0.73 & 0.77 \\
\bottomrule
\end{tabular}
\label{tab:rl_metrics}
\end{table*}

\subsection{Qualitative Evaluation \& Edge Inference}
\textbf{Mitigation of Degenerate Repetitive Loops:} Standard vision-language architectures are highly susceptible to representational enmeshment, leading to infinite generation loops. When evaluating the unaligned baseline on complex inputs (e.g., heavily swept delta wings), the generated \texttt{<think>} block collapses into repetitive cyclical text, triggering an \texttt{INVALID} regex extraction %
. GSPO thoroughly resolved this enmeshment; by evaluating sequence-level advantage, the policy efficiently synthesized RGB structural cues, thermal nacelle signatures, and LiDAR volume geometry into a concise deduction trail terminating in a correctly formatted cluster ID (e.g., \texttt{Answer:[C3]}).

\textbf{Systematic Error Dissection:} Despite achieving a highly balanced precision-recall matrix, GSPO encounters systematic failures at hardware resolution thresholds. The LiDAR projection depth limitations occasionally fail to resolve subtle stealth faceting (causing misclassifications between 5th-Gen and older Russian architectures) or mistake propeller hub thermal signatures for turbofan exhaust volumes.

\subsection{Hardware Optimization}
To validate edge viability, the aligned Qwen3-VL-4B-Instruct checkpoints are subjected to 4-bit normal-float quantization paired with attention-memory caching. This compression strictly bounds the framework within an 8 GB VRAM envelope. Latency profiling on edge-equivalent targets yields an end-to-end processing throughput of 1.48 seconds per multi-spectral triplet (an improvement from the 4.82-second FP16 baseline), establishing TriCLE as a highly competitive real-time classification system for resource-constrained platforms.

\subsection{Inference Generalization  \& Cross-Domain Evaluation}
\textbf{Aircraft Taxonomic Evaluation:}
To rigorously validate the real-world deployment viability of the TriCLE pipeline, the aligned checkpoints were evaluated against the unaligned foundational baseline on a strictly held-out test partition. This final inference stage utilizes 100 synchronized multi-spectral aircraft triplets, testing the models' capacity to synthesize novel optical, thermal, and LiDAR geometries without prior exposure to the specific airframes. The quantitative results, summarizing raw classification accuracy, format compliance (the successful generation of a parsable taxonomic tag), and F1-scores, are presented in Table \ref{tab:test_inference}. To ensure a rigorous evaluation against representational enmeshment, any outputs that collapsed into repetitive loops (yielding an \texttt{INVALID} extraction) were heavily penalized in the F1 calculations.

The empirical test data establishes the generalization superiority of the Group Sequence Policy Optimization (GSPO) framework. The GSPO-aligned checkpoint achieved a peak zero-shot accuracy of 78.00\% and a weighted F1-score of 0.793, successfully outperforming the unaligned baseline model (74.00\% accuracy). Furthermore, GSPO maintained a highly robust 94.00\% formatting validity rate. This confirms that sequence-level objective functions effectively prevent the structural degradation and context collapse that typically afflict parameter-efficient fine-tuning on edge-deployed VLMs. Conversely, policies reliant on token-level probability clipping (GRPO and DAPO) converged to identical, sub-optimal performance matrices on this test distribution. Both models recorded a diminished accuracy of 67.00\% and a pronounced drop in format compliance (85.00\%), indicating a distinct vulnerability to representational loops when processing unfamiliar multi-modal inputs. These test results definitively validate that sequence-level advantage estimation (GSPO) is the optimal alignment strategy for structurally rigorous aerospace classification tasks.

\begin{table}[h]
\centering
\caption{\footnotesize Zero-Shot Inference Performance on Held-Out Test Partition ($n=100$)}
\begin{tabular}{lcccc}
\toprule
\textbf{Model Pipeline} & \textbf{Test Accuracy} & \textbf{Format Valid Rate} & \textbf{Macro F1} & \textbf{Weighted F1} \\
\midrule
Unaligned Base Model & 74.00\% & 95.00\% & 0.647 & 0.752 \\
RPSFT + GRPO & 67.00\% & 85.00\% & 0.564 & 0.708 \\
RPSFT + DAPO & 67.00\% & 85.00\% & 0.564 & 0.708 \\
\textbf{RPSFT + GSPO (Ours)} & \textbf{78.00\%} & \textbf{94.00\%} & \textbf{0.654} & \textbf{0.793} \\
\bottomrule
\end{tabular}
\label{tab:test_inference}
\end{table}

\textbf{DSERT-RoLL Zero-Shot Generalization:}
While the preceding section analyzes our models' reasoning performance on the in-distribution aircraft domain, we now evaluate their cross-domain, zero-shot generalization capabilities. We deploy the frozen, aircraft-aligned policy checkpoints directly on the real-world, out-of-distribution road scenes of the DSERT-RoLL dataset.

Because hand-labeled, co-registered ground truths for multi-spectral road scenes are highly scarce, we implement an unsupervised VLM-as-a-Judge (Oracle Evaluation) framework. We employ the state-of-the-art, high-capacity \texttt{Qwen3-VL-32B-Thinking} model as our oracle evaluator to generate dense, multi-spectral silver ground truths over the co-registered image triplets. We then compare our model checkpoints against this oracle on both qualitative classifications and quantitative counting accuracies (Cars, Pedestrians, and Cycles). Table~\ref{tab:fused_dsert} details the empirical performance metrics.

\begin{table*}[t]
\centering
\resizebox{\textwidth}{!}{%
\begin{tabular}{lcccccccc}
\toprule
\textbf{Policy} & \textbf{Format Success} & \textbf{Exact Match} & \textbf{MAE Cars} & \textbf{RMSE Cars} & \textbf{MAE Peds} & \textbf{RMSE Peds} & \textbf{MAE Cyc} & \textbf{RMSE Cyc} \\
\midrule
RPSFT+GRPO & 100.00\% & 7.43\% & 1.5541 & 1.9779 & 0.5709 & 0.8562 & 0.4291 & 0.6550 \\
\textbf{RPSFT+GSPO} & \textbf{100.00\%} & \textbf{10.14\%} & 1.6419 & 2.1514 & \textbf{0.5439} & \textbf{0.8562} & \textbf{0.3885} & \textbf{0.6233} \\
RPSFT+DAPO & 100.00\% & 5.74\% & 1.6250 & 2.1077 & 0.5912 & 0.8872 & 0.4020 & 0.6341 \\
\bottomrule
\end{tabular}
}
\caption{\footnotesize Cross-domain zero-shot evaluation results on the co-registered DSERT-RoLL dataset graded against the \texttt{Qwen3-VL-32B-Thinking} oracle.}
\label{tab:fused_dsert}
\end{table*}

\textbf{Format Alignment Success.}
As shown in Table~\ref{tab:fused_dsert}, all three policy-aligned models (GRPO, GSPO, and DAPO) achieved 100.00\% Format Success Rate. This is a significant result: it demonstrates that our formatting reward ($R_f$) successfully trained the model's structural syntax to survive intense, out-of-distribution domain shifts. The models consistently generated complete reasoning traces inside the \texttt{<think>} blocks and ended with structured classifications and counts without a single syntax break.

\textbf{Qualitative Classification Analysis.}
The exact class match accuracy remains a challenging metric, with GSPO achieving the highest score at 10.14\%, followed by GRPO at 7.43\%, and DAPO at 5.74\%. The exact match metric is very strict, requiring that all vehicles, pedestrians and cycles in the scene be parsed with exact string match. Because our models were aligned on an aircraft taxonomy (with labels $C_0$ to $C_10$), attempting to classify 24 highly complex urban road objects under zero-shot conditions presents an extreme challenge.  Nevertheless, the better accuracy of the GSPO indicates the power of sequence-level trajectory optimization, which better preserves the global multi-spectral dependencies than token-level estimators, and reduces the visual attention drift in classifying multi-spectral features.

\textbf{Quantitative Counting Resilience.}
The models achieved very competitive performance on quantitative counting tasks. The Mean Absolute Error (MAE) for Cars was in the range of 1.55-1.64, meaning on average less than two vehicles per frame in heavy traffic. More importantly, the MAE for Pedestrians was also very low, being 0.54-0.59 (an average error of half a person), while the MAE for Cycles was 0.38-0.42. GSPO provided the lowest counting errors for both Pedestrians (MAE: 0.5439, RMSE: 0.8562) and Cycles (MAE: 0.3885, RMSE: 0.6233), demonstrating its ability to preserve and trace sparse and low-resolution visual tokens across co-registered inputs.

\section{Discussion}
\label{sec:discussion}

The results suggest that TriCLE's main gain comes from aligning multimodal reasoning with a structured taxonomic output, rather than from increasing model scale. All policy-alignment runs use the same compact 4B backbone, prompt format, and expert aircraft taxonomy, but they differ in how they update the model during long-form generation. GSPO gives the strongest overall behavior, reaching 88.33\% validation accuracy with 0.91 weighted F1, and maintaining the best held-out aircraft performance with 78.00\% accuracy, 0.793 weighted F1, and 94.00\% parseable outputs. This indicates that sequence-level policy optimization is better matched to the TriCLE setting, where the model must combine RGB appearance, thermal-style cues, and pseudo-LiDAR geometry before producing a valid cluster label. The comparison also shows that formatting validity alone is not enough. The RPSFT base model preserves structured outputs but gives weaker taxonomic accuracy, while token-level GRPO and DAPO are more vulnerable to accuracy and validity drops on unfamiliar tri-modal inputs. DAPO's overlong penalty helps control generation length, but in this task it can conflict with the required reasoning block before the final answer. GSPO avoids this trade-off more effectively by optimizing the completion at the sequence level, which helps preserve coherent reasoning without increasing the deployment footprint. The DSERT-RoLL experiment should be interpreted as a cross-domain stress test rather than direct aircraft validation. The aligned models preserve output syntax under a large domain shift, and GSPO obtains the strongest exact-match score among the tested policies. However, the road-scene targets are generated by a larger VLM evaluator, so the results measure silver-label agreement and structured-output robustness rather than human-verified perception accuracy. The held-out aircraft partition remains the primary evidence for TriCLE's target application.
Finally, the hardware profile supports the practical edge-deployment claim. With 4-bit quantization and attention-memory optimization, the aligned model fits within an 8GB VRAM target and processes each prepared tri-modal triplet in 1.48 seconds. This makes TriCLE a feasible prototype for edge-constrained aircraft grouping.
\vspace{-5pt}
\section{Limitations and Ethical Considerations}
\label{sec:limitations}

TriCLE remains a prototype system. Its thermal and pseudo-LiDAR inputs are synthetically derived from RGB images, so they may contain generator artifacts rather than physically measured heat or range. These synthetic modalities are useful for controlled training and stress testing, but they cannot replace calibrated thermal cameras, active LiDAR, or real sensor synchronization. Future work should evaluate TriCLE on co-registered RGB--thermal--LiDAR aircraft streams and compare RGB-only, synthetic tri-modal, real tri-modal, and mixed-modality inputs through explicit ablations. The evaluation is also limited by scale and coverage. The aircraft data are curated single-object images, and the held-out test partition contains 100 tri-modal triplets. This setting does not fully capture multi-aircraft scenes, which makes their per-class precision and recall less stable. A stronger benchmark should include hard negative aircraft pairs, unseen airframe families, low-resolution targets, and confidence-calibrated open-set rejection. The DSERT-RoLL evaluation has a separate limitation. It uses road scenes and VLM-generated silver labels, While useful for testing format robustness and cross-domain behavior but should not be treated as ground-truth validation for aircraft grouping. Human-labeled multispectral data are needed before real-world transfer.
\vspace{-5pt}
\section{Conclusion}\vspace{-5pt}
\label{sec:conclusion}

\textbf{TriCLE}, an edge-oriented tri-modal VLM system for aircraft taxonomic grouping. TriCLE fuses the three modalities with a compact Qwen3-VL backbone aligned to an engineering taxonomy. With 4-bit quantization, the model fits an 8GB edge target and processes each tri-modal triplet in 1.48 seconds. These results suggest that multi-sensor conditioning can support interpretable aircraft grouping under edge constraints, while future work should validate TriCLE on real co-registered sensor streams.

{\small
\bibliographystyle{ieeenat_fullname}
\bibliography{references}
}

\appendix

\section{Policy Optimization Implementation Specifics}

To support reproducibility, we report the exact training configuration used for each reinforcement learning alignment method. All three runs begin from the same RPSFT checkpoint, with all parameters frozen except the last 12 transformer layers. Within those layers, the attention projections (\texttt{q\_proj}, \texttt{k\_proj}, \texttt{v\_proj}, and \texttt{o\_proj}) are made trainable and cast to \texttt{float32}. The shared setup also keeps the dataset, prompt format, batch schedule, training budget, and reward function constant across methods so that the comparison isolates the effect of the policy optimization strategy itself. 
\begin{table}[!ht]
\centering
\resizebox{\columnwidth}{!}{%
\begin{tabular}{@{}llll@{}}
\toprule
\textbf{Configuration} & \textbf{GRPO} & \textbf{GSPO} & \textbf{DAPO} \\ \midrule
Importance sampling level & Token & Sequence & Token \\
Loss type & GRPO & GRPO with sequence-level IS & DAPO \\
Clip range & $\sim0.2/\sim0.2$ & $3{\times}10^{-4}/4{\times}10^{-4}$ & $0.2/0.28$ \\
Clipping style & Symmetric & Asymmetric & Asymmetric \\
KL penalty $\beta$ & $\sim0.04$ & $0.0$ & $\sim0.04$ \\
Mask truncated completions & No & No & Yes \\
Extra reward function & None & None & \texttt{soft\_overlong\_punishment} \\
\texttt{enable\_input\_require\_grads()} & No & Yes & Yes \\
Validation image pipeline & Processor direct & \texttt{process\_vision\_info} & \texttt{process\_vision\_info} \\ \bottomrule
\end{tabular}%
}
\vspace{1mm}
\caption{Implementation comparison of GRPO, GSPO, and DAPO alignment strategies.}
\label{tab:policy_impl}
\end{table}

\noindent Table~\ref{tab:policy_impl} summarizes the implementation differences among the three policy-alignment runs. GRPO serves as the token-level baseline with symmetric clipping and an active KL penalty, while GSPO replaces token-level clipping with sequence-level importance sampling, disables the KL penalty, and uses a much tighter asymmetric clip range. DAPO retains token-level optimization but adds asymmetric clipping, masks truncated completions, and applies a soft overlong punishment reward. All three methods use the same reward priorities: valid output formatting, correct taxonomic prediction, and coherent reasoning before the final answer. Since the task requires a full reasoning block before producing a parseable cluster label, methods that preserve long-form sequence coherence are favored. This explains why GSPO gives the most stable validation behavior, whereas DAPO is more sensitive to length penalties and produces fewer format-valid outputs.

\section{Per-Cluster Evaluation Results}

Table~\ref{tab:cluster_comparison} consolidates the per-cluster classification reports for GRPO, GSPO, and DAPO using valid answers only. Across the three alignment strategies, the strongest and most stable performance appears on better-supported or visually distinctive clusters, especially C3, C6, C7, and C9. In contrast, low-support clusters such as C5 and C10 show higher variance, so their scores should be interpreted cautiously.

GRPO provides stable performance across most clusters, with strong weighted results and consistent behavior on the dominant classes. GSPO gives the strongest macro-level balance and maintains high precision on several difficult clusters, suggesting that sequence-level optimization helps preserve coherent multimodal reasoning when integrating RGB, thermal, and LiDAR cues. DAPO remains competitive on C3, C6, C7, and C9, but it shows stronger precision--recall asymmetry on C1 and C4. This pattern is consistent with token-level dynamic clipping producing uneven class-level effects, especially when the valid-answer pool is smaller and the class support is limited.

\begin{table*}[h]
\centering
\renewcommand{\arraystretch}{0.90}
\begin{tabular}{@{}lrrrrrrrrrrrr@{}}
\toprule
& \multicolumn{4}{c}{\textbf{GRPO}} 
& \multicolumn{4}{c}{\textbf{GSPO}} 
& \multicolumn{4}{c}{\textbf{DAPO}} \\
\cmidrule(lr){2-5}
\cmidrule(lr){6-9}
\cmidrule(l){10-13}
\textbf{Cluster} 
& \textbf{P} & \textbf{R} & \textbf{F1} & \textbf{n}
& \textbf{P} & \textbf{R} & \textbf{F1} & \textbf{n}
& \textbf{P} & \textbf{R} & \textbf{F1} & \textbf{n} \\
\midrule
C1  & 0.73 & 0.92 & 0.81 & 15 & 1.00 & 0.87 & 0.93 & 15 & 0.89 & 0.57 & 0.70 & 14 \\
C2  & 0.81 & 0.87 & 0.84 & 16 & 0.92 & 0.80 & 0.86 & 15 & 0.89 & 0.80 & 0.84 & 10 \\
C3  & 0.96 & 0.96 & 0.96 & 28 & 0.96 & 0.98 & 0.97 & 27 & 0.93 & 0.96 & 0.94 & 26 \\
C4  & 1.00 & 0.70 & 0.82 & 7  & 0.89 & 0.94 & 0.91 & 8  & 0.60 & 1.00 & 0.75 & 6  \\
C5  & 0.67 & 0.67 & 0.67 & 3  & 0.67 & 0.80 & 0.73 & 2  & 0.50 & 0.67 & 0.57 & 3  \\
C6  & 0.96 & 0.85 & 0.90 & 23 & 0.85 & 0.79 & 0.82 & 25 & 0.83 & 0.95 & 0.89 & 21 \\
C7  & 1.00 & 1.00 & 1.00 & 5  & 1.00 & 1.00 & 1.00 & 5  & 1.00 & 1.00 & 1.00 & 4  \\
C8  & 1.00 & 0.80 & 0.89 & 9  & 0.80 & 1.00 & 0.89 & 10 & 1.00 & 0.75 & 0.86 & 8  \\
C9  & 0.83 & 1.00 & 0.91 & 5  & 0.83 & 1.00 & 0.91 & 5  & 1.00 & 1.00 & 1.00 & 4  \\
C10 & 0.60 & 1.00 & 0.75 & 5  & 1.00 & 0.60 & 0.75 & 5  & 1.00 & 0.67 & 0.80 & 3  \\
\midrule
\textbf{Macro Avg}    
& \textbf{0.86} & \textbf{0.88} & \textbf{0.86} & \textbf{116}
& \textbf{0.89} & \textbf{0.88} & \textbf{0.88} & \textbf{117}
& \textbf{0.86} & \textbf{0.84} & \textbf{0.83} & \textbf{99} \\
\textbf{Weighted Avg} 
& \textbf{0.90} & \textbf{0.89} & \textbf{0.89} & \textbf{116}
& \textbf{0.91} & \textbf{0.88} & \textbf{0.89} & \textbf{117}
& \textbf{0.88} & \textbf{0.86} & \textbf{0.86} & \textbf{99} \\
\bottomrule
\end{tabular}
\vspace{1mm}
\caption{\footnotesize Merged per-cluster classification reports for GRPO, GSPO, and DAPO, computed on valid answers only. Here, P, R, F1, and $n$ denote precision, recall, F1-score, and valid-answer support, respectively.}
\label{tab:cluster_comparison}
\end{table*}

\begin{table*}[t]
\centering
\renewcommand{\arraystretch}{0.92}
\begin{tabularx}{\columnwidth}{|>{\raggedright\arraybackslash}p{0.27\columnwidth}|Y|Y|Y|}
\hline
\textbf{Metric} & \textbf{GRPO} & \textbf{GSPO} & \textbf{DAPO} \\
\hline
Family & Token GRPO & Sequence GSPO & Token DAPO \\
\hline
Importance sampling & Token & Sequence & Token \\
\hline
Clip bounds L/U & $\sim0.2/\sim0.2$ & $3{\times}10^{-4}/4{\times}10^{-4}$ & $0.2/0.28$ \\
\hline
KL penalty $\beta$ & $\sim0.04$ & $0.0$ & $\sim0.04$ \\
\hline
Truncated mask & No & No & Yes \\
\hline
Extra reward & None & None & \shortstack[c]{\texttt{soft\_overlong}\\\texttt{punishment}} \\
\hline
Learning rate & $1{\times}10^{-5}$ & $1{\times}10^{-5}$ & $1{\times}10^{-5}$ \\
\hline
Eff. batch size & 16 & 16 & 16 \\
\hline
Max length & 768 & 768 & 768 \\
\hline
Training time & \shortstack[c]{$\sim$7h 44m\\27,864 s} & \shortstack[c]{$\sim$7h 25m\\26,699 s} & \shortstack[c]{$\sim$7h 9m\\25,765 s} \\
\hline
Format valid & \shortstack[c]{4,710/4,800\\98.12\%} & \shortstack[c]{4,704/4,800\\98.00\%} & \shortstack[c]{4,586/4,800\\95.54\%} \\
\hline
Training acc. & \shortstack[c]{3,867/4,710\\82.10\%} & \shortstack[c]{3,913/4,704\\83.18\%} & \shortstack[c]{3,719/4,586\\81.09\%} \\
\hline
Val. acc. $(n=120)$ & 85.83\% & \textbf{88.33\%} & 75.83\% \\
\hline
Weighted F1 & \shortstack[c]{0.89\\$(n=116)$} & \shortstack[c]{\textbf{0.91}\\$(n=117)$} & \shortstack[c]{0.86\\$(n=99)$} \\
\hline
\end{tabularx}
\vspace{1mm}
\caption{\footnotesize One-column comparison of GRPO, GSPO, and DAPO alignment strategies.}
\label{tab:cross_method_comparison}
\end{table*}

\begin{figure}[!hbp]
    \centering
    \includegraphics[width=0.7\linewidth]{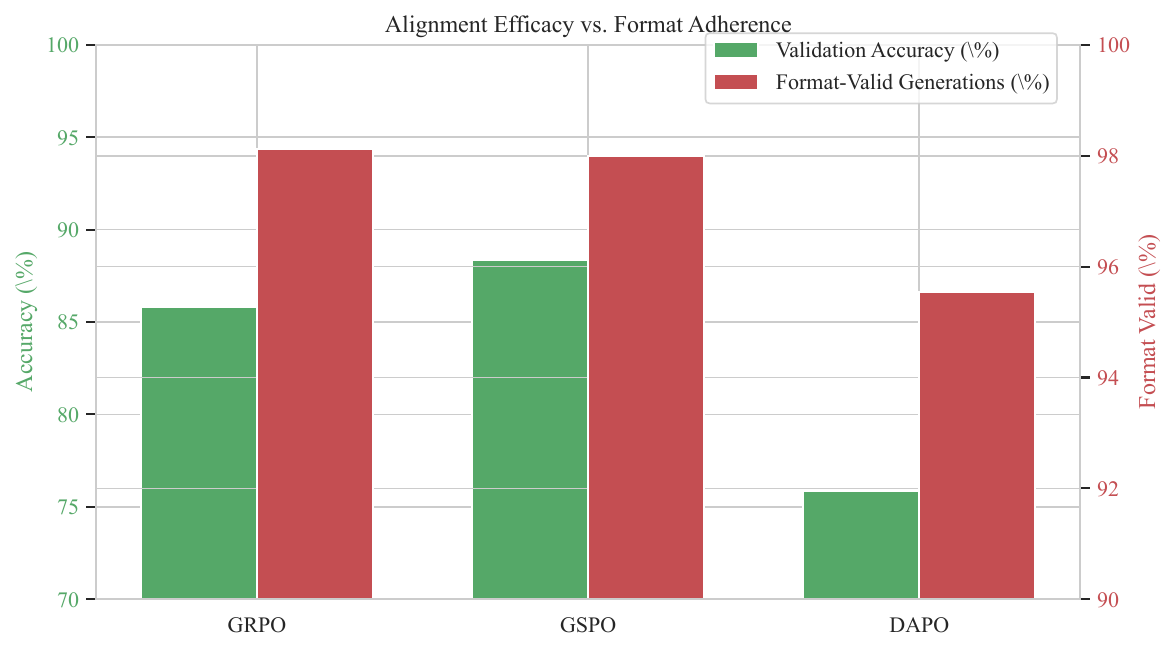}
    \caption{\textbf{Effectiveness of alignment vs.\ compliance with format.} Validity accuracy (left axis, green) and format-valid generation percentage (right axis, red) for three alignment strategies. Aggressive clipping reduces training duration but lowers reasoning accuracy and format. GSPO maintains near-perfect format validity (98.00\%) and optimal validation accuracy, proving sequence-level optimisation does not compromise instruction adherence.}
    \label{fig:acc_format}
\end{figure}

\begin{figure}[!hbp]
    \centering
    \includegraphics[width=0.7\linewidth]{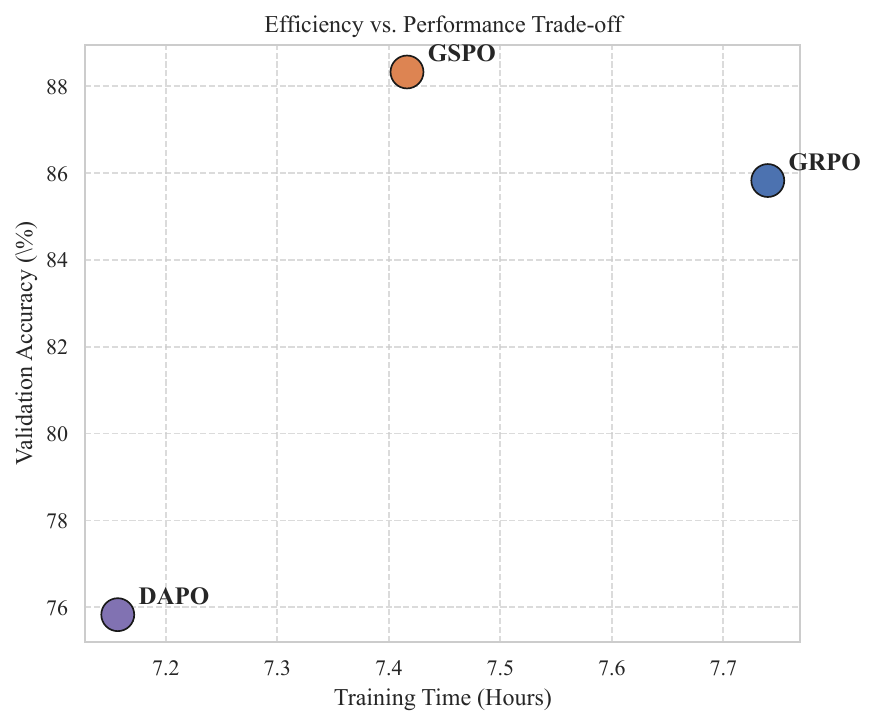}
    \caption{\textbf{Computational efficiency vs. performance.} Validation accuracy plotted against total training time for GRPO, GSPO, and DAPO. GSPO falls within the Pareto-optimal range, achieving 88.33\% validation accuracy with 2\% less training time than token-level GRPO. This shows that sequence-level importance sampling improves performance without increasing computational overhead.}
    \label{fig:efficiency}
\end{figure}

\section{Cross-Method Comparison}

Table~\ref{tab:cross_method_comparison} consolidates the main differences between the three alignment strategies. The table highlights that all methods share the same base model, training budget, and output format, but differ in their clipping rules, sampling level, validation preprocessing, and auxiliary reward design. This makes the comparison useful as a controlled ablation of optimization strategy rather than a comparison of unrelated pipelines.


\noindent Overall, the appendix supports three clear conclusions. First, the RPSFT checkpoint provides a stable and reusable starting point for all RL methods. Second, GSPO offers the best balance of stability and performance for this long-form reasoning task. Third, DAPO is less suitable here because the overlong penalty conflicts with the deliberate reasoning structure required by the prompt.
\end{document}